\documentclass[runningheads]{llncs}
\usepackage{graphicx}
\usepackage{amsmath,amssymb} 
\usepackage{color}

\usepackage{algorithm,algpseudocode}

\begin{document}
\pagestyle{headings}
\mainmatter

\def\ACCV20SubNumber{123}  

\title{Learning End-to-End Action Interaction by Paired-Embedding Data Augmentation} 
\titlerunning{ACCV-20 submission ID \ACCV20SubNumber}
\authorrunning{ACCV-20 submission ID \ACCV20SubNumber}

\author{  Ziyang Song$^1$, Zejian Yuan$^1$, Chong Zhang$^2$, Wanchao Chi$^2$, Yonggen Ling$^2$, and Shenghao Zhang$^2$ \\  \thanks{Correspondance: {\tt\small songzy305@yahoo.com}}
}
\institute{
$^1$ Institute of Artificial Intelligence and Robotics, Xi'an Jiaotong University, Xi'an, China\\  
$^2$ Tencent Robotics X, Shenzhen, China\\
}

\maketitle

\begin{abstract}
In recognition-based action interaction, robots' responses to human actions are often pre-designed according to recognized categories and thus stiff.
In this paper, we specify a new Interactive Action Translation (IAT) task which aims to learn end-to-end action interaction from unlabeled interactive pairs, removing explicit action recognition.
To enable learning on small-scale data, we propose a Paired-Embedding (PE) method for effective and reliable data augmentation.
Specifically, our method first utilizes paired relationships to cluster individual actions in an embedding space.
Then two actions originally paired can be replaced with other actions in their respective neighborhood, assembling into new pairs.
An Act2Act network based on conditional GAN follows to learn from augmented data.
Besides, IAT-test and IAT-train scores are specifically proposed for evaluating methods on our task.
Experimental results on two datasets show impressive effects and broad application prospects of our method.
\end{abstract}

\section{Introduction}

Action interaction is an essential part of human-robot interaction (HRI)~\cite{HRI}.
For robots, action interaction with human includes two levels:
1) perceiving human actions and understanding intentions behind;
2) performing responsive actions accordingly.
Thanks to the development of action recognition methods~\cite{survey_actionHRI}, considerable progress has been made on the first level.
As for the second level, robots often perform pre-designed action responses according to recognition results.
We call this scheme as recognition-based action interaction.
However, colorful appearances of human actions are mapped to a few fixed categories in this way, leading to a few fixed responses.
Robots' action responses are thus stiff, lacking in human-like vividity.
Moreover, annotating data for training action recognition models consumes manpower.

In this paper, we aim to learn end-to-end interaction from unlabeled action interaction data.
Explicit recognition is removed, leaving the interaction implicitly guided by high-level semantic translation relationships.
To achieve this goal, we specify a novel Interactive Action Translation (IAT) task:
Given a set of "stimulation-response" action pairs conforming to defined interaction rules and without category labeled, learn a model to generate a response for a given stimulation during inference.
The generated results are expected to manifest:

\textbf{1) reality:} indistinguishable from real human actions;

\textbf{2) precision:} conforming to defined interaction rules semantically, conditioned on the stimulation;

\textbf{3) diversity:} be various each time given the same stimulation.

For different interaction scenes and defined rules, paired action data need to be re-collected each time.
Thus IAT would be more appealing if learning from a small number of samples.
However, the task implicitly seeks for a high-level semantic translation relationship, which is hard to generalize from insufficient data.
Moreover, the multimodal distribution of real actions is difficult to approximate without sufficient data.
The contradiction between task goals and applications poses the main challenge: to achieve the three generation goals above with small-scale data.

Data augmentation is widely adopted to improve learning on small datasets.
Traditional augmentation strategies apply hand-crafted transformations on existing data, thus only bring changes in limited modes.
Generative Adversarial Networks (GAN)~\cite{GAN} emerges as a powerful technique to generate realistic samples.
Nonetheless, a reliable GAN itself requires large-scale data to train.
Some variants of GAN, like ACGAN~\cite{ACGAN}, DAGAN~\cite{DAGAN}, and BAGAN~\cite{BAGAN}, are proposed to augment data for classification tasks.
However, all of them need category labels that are not provided in our task.
Therefore, a specially designed augmentation method is needed for small-scale unlabeled data in IAT.

\begin{figure}[t]
\centering
\includegraphics[width=105mm]{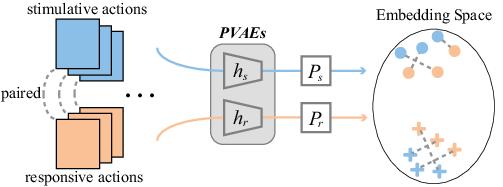}
\caption{
An overview of our proposed Paired-Embedding (PE) method. 
Colors distinguish stimulations and responses. 
Circle and cross denote actions of different semantic categories.
Dotted lines describe paired relationships.
}
\label{fig:overview}
\end{figure}

We propose a novel Paired-Embedding (PE) method, as Fig.~\ref{fig:overview} shows.
Through encoders in a Paired Variational Auto-Encoders (PVAEs) and PCA-based linear dimension reductions, individual action instances are projected into a low-dimension embedding space.
Along with the vanilla VAE objectives~\cite{VAE}, we employ a new PE loss utilizing paired relationships between actions to train PVAEs.
Specifically, VAE loss prefers large variance of action embeddings while PE loss pull actions within the same categories together.
As a result, action instances are clustered in the embedding space in an unsupervised manner.
Subsequently, both two actions in a data pair are allowed to be replaced with other instances in their respective neighborhood, assembling into new pairs conforming to defined interaction rules semantically.
Therefore, the diversity of paired data is significantly and reliably enriched.
Finally, we train an Act2Act network based on conditional GAN~\cite{cGAN} on augmented data to solve our task.

Although IAT is formally an instance-conditional generation task like image translation~\cite{pix2pix,BicycleGAN}, it actually conditions on the semantic category of input action instances.
Therefore, evaluation metrics for neither image translation~\cite{AMT,LPIPS} nor category-conditional generation~\cite{GAN-train&test} is suitable for this task.
Considering the three generation goals, we propose two evaluation metrics, IAT-test and IAT-train scores, to compare methods for our task from distinct perspectives.
Experiments show that our proposed method gives satisfying generated action responses, both quantitatively and qualitatively.

The major contributions of our work are summarized as follows:

1) We specify a new IAT task, aiming to learn end-to-end action interaction from unlabeled interactive action pairs.

2) We design a PE data augmentation method to resolve the main challenge of our task: learning with a small number of samples.

3) We propose IAT-test and IAT-train scores to evaluate methods on our task, covering three task goals.
Experiments prove the satisfying generation effects of our proposed method.

\section{Related Work}

\subsection{Data Augmentation with GAN}

It is widely accepted that in deep learning, a larger dataset often leads to a more reliable algorithm.
In practical applications, data augmentation by adding synthetic data provides another way to improve performance.
The most common data augmentation strategies are applying various hand-designed transformations on existing data.
As GAN arises, it is a straightforward idea to use GAN to directly synthesize realistic data for augmentation.
However, GAN itself always requires large-scale data for stable training.
Otherwise, the quality of synthesized data is not ensured.

Several variants of conditional GAN are proposed for augmenting classification tasks, where category labels are included in GAN training.
ACGAN~\cite{ACGAN} lets the generator and discriminator 'cooperating' on classification in addition to 'competing' on generation.
DAGAN~\cite{DAGAN} aims to learn transformations on existing data for data augmentation.
BAGAN~\cite{BAGAN} restores the dataset balance by generating minority-class samples.
Unfortunately, these methods can not be applied to augmenting data without category labels given.
Some other GAN-based data augmentation methods are also designed for different tasks, like~\cite{GANaug_emc} for emotion classification and~\cite{GANaug_reid1,GANaug_reid2} for person re-identification.
They are only suitable for respective tasks but not extensible to our task.
Unlike these methods, our proposed method augments IAT data by re-assigning individual actions from existing pairs into new pairs. 
Data synthesized in this way are undoubtedly natural and realistic.
Meanwhile, PE method ensures the same interaction rules on augmented data and existing data, namely the semantic-level reality of augmented data.

\subsection{Evaluation Metrics for Generation}

Early work often relies on subjective visual evaluation of synthesized samples from generative methods like GAN.
Quantitative metrics are proposed in recent years, and the most popular among them are Inception score (IS)~\cite{IS} and Fr\'{e}chet Inception distance (FID)~\cite{FID}.
Both of them are based on a pre-trained classification network (for image generation, an Inception network pre-trained on ImageNet).
IS predicts category probabilities on generated samples through the classification network and evaluates generated results accordingly.
FID directly measures the divergence between distributions of real and synthesized data in feature-level.
CSGN~\cite{CSGN} has extended IS and FID metrics from image generation to skeleton-based action synthesis.
However, they fail to reflect the dependence of generated results upon conditions, thus are unsuitable for conditional generation tasks like ours.

GAN-train and GAN-test scores~\cite{GAN-train&test} are proposed for comparing category-conditional GANs.
An additional classification network is also introduced.
Given category information, the two metrics quantify the correlation between generated samples and conditioned categories besides generating reality and diversity.
Nonetheless, category labels are missing in our task and semantic categories are implicitly reflected in paired relationships.
Enlightened by GAN-train and GAN-test, we propose IAT-test and IAT-train scores to fit our task.
In our metrics, binary classification on data pairs is adopted in the classification network instead of explicit multi-category classification on individual instances.

\section{Proposed Method}

Our method consists of two parts: a core Paired-Embedding (PE) method for effective and reliable data augmentation, and an Act2Act network following the former.
We illustrate the two parts separately in the following.

\subsection{Paired-Embedding Data Augmentation}

\begin{figure}[t]
\centering
\includegraphics[width=120mm]{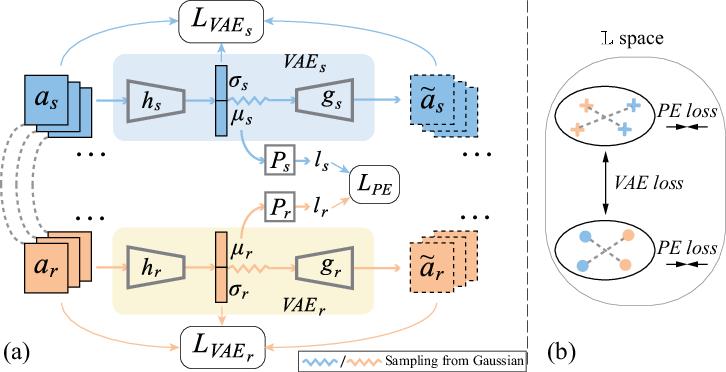}
\caption{
(a) The structure of Paired Variational Auto-Encoders (PVAEs) and losses for training.
(b) Effects of different losses.
}
\label{fig:pecluster}
\end{figure}

Here we propose a Paired-Embedding (PE) method, which aims to cluster individual action instances in a low-dimension embedding space by utilizing paired relationships between them.

\subsubsection{Paired Variational Auto-Encoders (PVAEs).}

PE is based on a Paired Variational Auto-Encoders (PVAEs) consisting of two separate Variational Auto-Encoder (VAE)~\cite{VAE} networks $VAE_{s}$ and $VAE_{r}$ with the same architecture, as shown in Fig.~\ref{fig:pecluster}(a).
Following~\cite{VAE}, a VAE network is composed of an encoder $h$ and a decoder $g$.
The encoder projects each sample $a$ into ($\mu$, $\sigma$), which are parameters of a multivariate Gaussian distribution $N(\mu, \sigma^2 I)$.
Then a latent variable is sampled from this distribution to generate $\widetilde{a}$ through the decoder.
Reconstruction error from $\widetilde{a}$ to $a$ and a prior regularization term constitutes VAE loss, i.e.,
\begin{equation}
    L_{VAE} (a, \widetilde{a}, \mu, \sigma) = || a - \widetilde{a} || ^ 2 + \lambda_{KL} D_{KL} ( N(\mu, \sigma^2 I) || N(0, I))
\end{equation}
where $D_{KL}$ is the Kullback-Leibler divergence, with $\lambda_{KL}$ controlling its relative importance.

We extract individual action instances from original action pairs.
The two networks can be respectively trained under VAE loss to model the distribution of stimulative/responsive actions.

\subsubsection{Paired-Embedding (PE) Loss.}

Given an action set, the encoder of VAE projects each action into a $\mu$ as the mean of a Gaussian distribution.
We collect Gaussian means from all the actions and compute a matrix $P$ for linear dimension reduction, using Principal Component Analysis (PCA) on them.
These Gaussian means are further projected by $P$ into an extremely low-dimension embedding space $\mathbb{L}$, namely as $l = P \mu$.
Owing to PCA, the variance of Gaussian means is well maintained in the $\mathbb{L}$ space.
Both stimulative and responsive actions are projected into the embedding space in this way.
For two actions paired in the original dataset $\rm \textbf{A}$, we push them towards each other in the embedding space using a Paired-Embedding (PE) loss, i.e.,
\begin{equation}
    L_{PE}(l_s, l_r) = || l_{s} - l_{r} || ^ 2 ,
\end{equation}
where $l_{s}$ and $l_{r}$ are embeddings of an interactive pair of actions in the $\mathbb{L}$ space.
Fig.~\ref{fig:pecluster}(a) illustrates such a process.

\subsubsection{Training PVAEs.}

We train $VAE_{s}$ and $VAE_{r}$ synchronously and divide each epoch into two steps, as in Algorithm~\ref{alg:train_pvae}.
During the first step, the two networks are independently optimized towards minimizing respective VAE loss.
In the second step, PE loss serves to guide encoders in two networks.

\begin{algorithm}[t]
    \caption{Training of PVAEs}
    \label{alg:train_pvae}
    \begin{algorithmic}[1]
        \Require {${\rm \textbf{A}} = \{ \dots, (a_s, a_r), \dots \}$}
        \Ensure {$h_s$, $g_s$, $h_r$, $g_r$}
        \State {Initialize $h_s$, $g_s$, $h_r$, $g_r$}
        \For {$epoch$ in [1, $Epochs$]}
            \State {\textcolor{blue}{\# First step under VAE loss}}
            \State {$L_{VAE_s} = 0$, $L_{VAE_r} = 0$}
            \For {$(a_s, a_r)$ in $\rm \textbf{A}$}
                \State {$(\mu_s, \sigma_s) = h_s (a_s)$, $(\mu_r, \sigma_r) = h_r (a_r)$}
                \State {Sample $z_s \sim N(\mu_s, {\sigma_s}^2 I)$, Sample $z_r \sim N(\mu_r, {\sigma_r}^2 I)$}
                \State {$\widetilde{a_s} = g_s (z_s)$,  $\widetilde{a_r} = g_r (z_r)$}
                \State {$L_{VAE_s}$ += $L_{VAE} (a_s, \widetilde{a_s}, \mu_s, \sigma_s)$, $L_{VAE_r}$ += $L_{VAE} (a_r, \widetilde{a_r}, \mu_r, \sigma_r)$}
            \EndFor
            \State {Back-prop $L_{VAE_s}$, update $h_s$, $g_s$; Back-prop $L_{VAE_r}$, update $h_r$, $g_r$}
            \State {\textcolor{blue}{\# Second step under PE loss}}
            \State {${\rm \textbf{M}}_s = \{ \}$, ${\rm \textbf{M}}_r = \{ \}$, ${\rm \textbf{M}} = \{ \}$, $L_{P} = 0$}
            \For {$(a_s, a_r)$ in $\rm \textbf{A}$}
                \State {$(\mu_s, \sigma_s) = h_s (a_s)$, $(\mu_r, \sigma_r) = h_r (a_r)$}
                \State {${\rm \textbf{M}}_s$.append($\mu_s$), ${\rm \textbf{M}}_r$.append($\mu_r$), ${\rm \textbf{M}}$.append($(\mu_s, \mu_r)$)}
            \EndFor
            \State {$P_s$ = PCA(${\rm \textbf{M}}_s$), $P_r$ = PCA(${\rm \textbf{M}}_r$)}
            \For {$(\mu_s, \mu_r)$ in $\rm \textbf{M}$}
                \State {$l_s = P_s \mu_s$, $l_r = P_r \mu_r$}
                \State {$L_P$ += $L_{PE} (l_s, l_r)$}
            \EndFor
            \State {Back-prop $L_P$, update $h_s$, $h_r$}
        \EndFor
    \end{algorithmic}
\end{algorithm}

Such an alternating strategy drives PVAEs from two opposite directions, as Fig.~\ref{fig:pecluster}(b) shows.
\begin{itemize}
    \item
    On the one hand, Gaussian means should scatter for the reconstruction of different action instances.
    In other words, Gaussian means must maintain a sufficiently large variance, which is transfered almost losslessly to $\mathbb{L}$ space by PCA.
    Consequently, the first learning step under VAE loss requires a large variance among $\mathbb{L}$ embeddings of stimulative/responsive actions respectively.
    \item
    On the other hand, each defined interaction rule is shared among several action pairs.
    For these action pairs, semantic category information is unified while other patterns in action instances are diverse.
    Since $\mathbb{L}$ space has an extremely low dimension, embeddings of paired actions can not be close for all pairs if the space mostly represents patterns apart from semantics.
    In other words, PE loss pushes the space towards representing semantic categories of actions only.
    Thus, stimulative or responsive actions within the same semantic category are pulled together in $\mathbb{L}$ space, guided by PE loss.
\end{itemize}

As a result, actions with similar semantics tend to cluster in the embedding space.
Meanwhile, different clusters are far away from each other to maintain large variance.
Experimental results in Sec. 4.4 further verify this effect.

\subsubsection{Data Augmentation with PVAEs.}

Given a set of individual action instances (either stimulative or responsive) and the corresponding VAE network from trained PVAEs, an $N \times N$ matrix $C$ is computed as,
\begin{equation}
    C(i, j) = exp( - \frac{|| l^{(i)} - l^{(j)} || ^ 2}{2 ||s \cdot (P \sigma^{(i)})|| ^ 2} ) ,
\end{equation}
where $N$ is the number of action instances, with $i$ and $j$ indexing two samples.
A pre-set scale factor $s$ controls the neighborhood range.
After that, we normalize the sum of each row in $C$ to 1, i.e.,
\begin{equation}
    NC(i, j) = \frac{C(i, j)}{\sum_{k=1}^N C(i, k)} .
\end{equation}

The computed $NC$ matrix represents confidence in replacing one action with another under defined interaction rules.
An action is believed to express semantics similar to other actions in its neighborhood, owing to clustering effects in $\mathbb{L}$ space.
We respectively compute two $NC$ matrices for stimulative and responsive action instances and use them to augment action pairs.
Two actions from each action pair in the original dataset are replaced with other samples in their respective neighborhood, according to $NC$ matrices.
Assume that $N$ data pairs in the original set are evenly distributed in $K$ semantic categories.
With replacement, we can optimally attain $\frac{N}{K} \times \frac{N}{K} \times K =\frac{N^2}{K}$ various data pairs conforming to defined interaction rules.
Such an increase in data diversity will significantly boost the learning effects of IAT task.

\subsection{Act2Act: Encoder-Deoder Network under Conditional GAN}

IAT is similar to paired image translation in the task form and goals.
Both of them can be regarded as an instance-conditional generation task.
They differ in that image translation conditions on the structured content of input instance, while our task implicitly conditions on the higher-level semantics of input instance.
In recent years, GAN-based methods have been successful in image translation, generating photorealistic results.
A similar GAN-based scheme is applied to our task.

\begin{figure}[t]
\centering
\includegraphics[width=120mm]{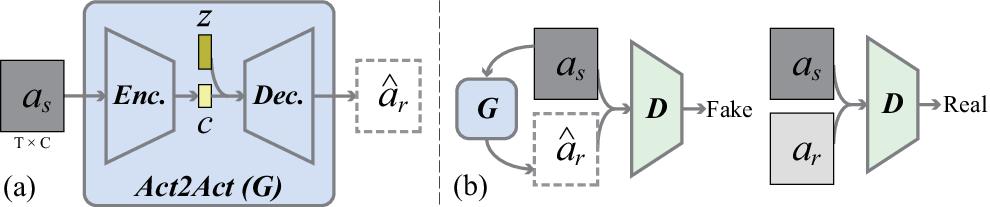}
\caption{
(a) The Act2Act network and (b) training under conditional GAN.
}
\label{fig:act2act}
\end{figure}

Our Act2Act network is stacked with an encoder-decoder architecture, as in Fig.~\ref{fig:act2act}(a).
It receives a stimulative action $a_s$ as input, and gives an output $\widehat{a}_{r}$ with the same form.
Through the encoder, a low-dimension code $c$ is extracted from $a_{s}$.
A random noise vector $z$ is sampled from zero-mean Gaussian distribution with unit variance, and then combined with $c$ to decode $\widehat{a}_{r}$.

Conditional GAN is applied for training, as Fig.~\ref{fig:act2act}(b) shows.
The encoder-decoder network is treated as Generator $G$, with another Discriminator $D$ receives a combination of two action sequences and outputs a score.
Given paired training data $(a_{s}, a_{r})$, $G$ is trained to produce $\widehat{a}_{r}$ indistinguishable from $a_{r}$.
Meanwhile, $D$ is trained to differentiate $(a_{s}, \widehat{a}_{r})$ from $(a_{s}, a_{r})$ as well as possible.

Behind the above design lies our understanding of IAT task.
We consider the task as an implicit series connection of recognition and category-conditional generation.
Therefore, we do not introduce $z$ until input is extracted into $c$, unlike in \cite{pix2pix,BicycleGAN} for image translation.
The code $c$ has a very low dimension since we expect it to encode high-level semantics.
Correlation between $a_{s}$ and $a_{r}$ exists only in semantics, but not low-level appearance.
Thus the encoder-decoder network is supervised by conditional GAN only, without reconstruction error from $\widehat{a}_{r}$ to $a_{r}$.
 
\section{Experiments}

\subsection{IAT-test and IAT-train}

\begin{figure}[t]
\centering
\includegraphics[width=110mm]{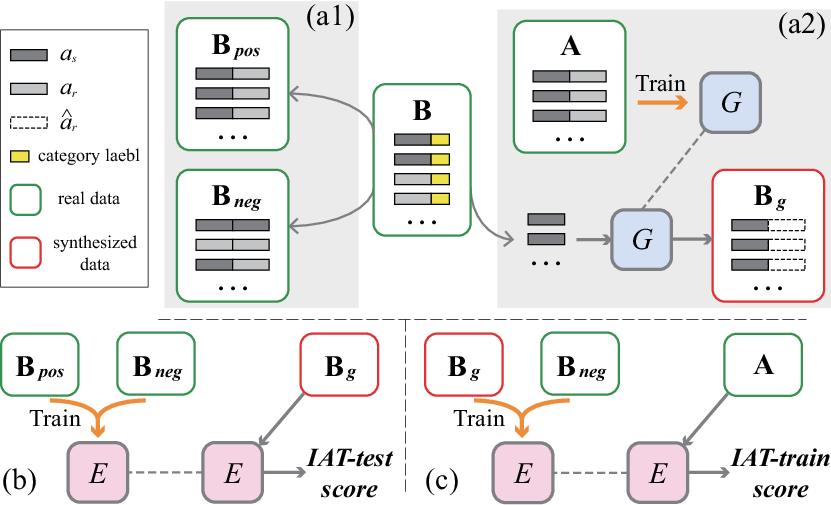}
\caption{
Illustration of our proposed evaluation metrics.
}
\label{fig:metric}
\end{figure}

Inspired by \cite{GAN-train&test}, we propose IAT-test and IAT-train scores to evaluate methods on our task, as illustrated in Fig.~\ref{fig:metric}.
Besides the training set ${\rm \textbf{A}}$ for the task, another set ${\rm \textbf{B}}$ composed of individual actions is introduced.
Categories of actions in set ${\rm \textbf{B}}$ are annotated.
Based on annotations, we can pair actions in ${\rm \textbf{B}}$ and assign pairs to ${\rm \textbf{B}}_{pos}$ or ${\rm \textbf{B}}_{neg}$. 
The former contains action pairs under the same interaction rules as $A$, while the latter contains the rest, as Fig.~\ref{fig:metric}(a1) shows.
Given a model $G$ trained on set ${\rm \textbf{A}}$, we select stimulative actions from ${\rm \textbf{B}}$ and generate responses for them, resulting in paired action set ${\rm \textbf{B}}_g$.
Fig.~\ref{fig:metric}(a2) illustrates such a process.
We evaluate the model $G$ according to ${\rm \textbf{B}}_g$ samples in the following ways.

\subsubsection{IAT-test.}

With positive samples from ${\rm \textbf{B}}_{pos}$ and negative samples from ${\rm \textbf{B}}_{neg}$, we train a binary classifier $E$ to judge whether an action pair accords to the defined interaction rules and give a 1/0 score accordingly.
K-fold cross-validation is adopted to investigate and ensure the generalization performance of $E$.

IAT-test is the test score of model $E$ on set ${\rm \textbf{B}}_g$, as shown in Fig.~\ref{fig:metric}(b).
If ${\rm \textbf{B}}_g$ is provided by a perfect model $G$, IAT-test score should approximate the K-fold validation accuracy of model $E$ during training.
Otherwise, a lower score can be attributed to:
1) Generated responses are not realistic enough;
2) Semantic translation relationships captured by $G$ are not precise, especially when generalized to stimulative actions in set ${\rm \textbf{B}}$.
In other words, IAT-test quantifies how well the generation goals of reality and precision are achieved.

\subsubsection{IAT-train.}

Here a classifier $E$ similar to the above is trained, with positive samples from ${\rm \textbf{B}}_g$ and negative samples from ${\rm \textbf{B}}_{neg}$.

IAT-train is the test score of model $E$ on set ${\rm \textbf{A}}$, as shown in Fig.~\ref{fig:metric}(c).
A low score can appear due to:
1) From unrealistic generation results, $E$ learns features useless for classifying real samples;
2) Incorrect interaction relationships in ${\rm \textbf{B}}_g$ misleads the model $E$.
3) Lack of diversity in ${\rm \textbf{B}}_g$ impairs the generalization performance of $E$.
Overall, IAT-train reflects the achievement of all three goals.

Combining the two metrics helps separate diversity from the other generation goals.
In other words, when the model $G$ receives a high IAT-test score and a low IAT-train score, the latter can be reasonably attributed to a poor generation diversity.

\subsection{Dataset}

We evaluate our method on UTD-MHAD~\cite{UTD_MHAD} and AID~\cite{AID} datasets, both composed of skeleton-based single-person daily interactive actions.
For each dataset, action categories are firstly paired to form our defined interaction rules, such as "tennis serve - tennis swing", "throw - catch", etc.
Then action clips in the dataset are divided into two parts:
clips in one part are randomly paired according to interaction rules to form set ${\rm \textbf{A}}$ for learning our task;
clips in the other part are reserved as set ${\rm \textbf{B}}$ for evaluation.

\subsubsection{UTD-MHAD}

consists of 861 action clips from 27 categories performed by 8 subjects.
Each frame describes a 3D human pose with 20 joints of the whole body.
We select 10 of 27 action categories and pair them into 5 meaningful interaction rules.
Moreover, we choose to use 9 joints of the upper body only since other joints hardly participate in selected actions.
Finally, we obtain a set ${\rm \textbf{A}}$ of 80 action pairs and a set ${\rm \textbf{B}}$ of 160 individual action instances.

\subsubsection{AID}

consists of 102 long sequences, each containing several short action clips.
Each frame describes a 3D human pose with 10 joints of the upper body.
After removing 5 corrupted sequences, we have 97 sequences left, performed by 19 subjects and covering 10 action categories.
Subsequently, 5 interaction rules are defined on the 10 categories.
Finally, we obtain a set ${\rm \textbf{A}}$ of 282 action pairs and a set ${\rm \textbf{B}}$ of 407 individual action instances.

\subsubsection{Implementation Details.}

Similar to~\cite{text2action}, action data are represented as normalized limb vectors instead of original joint coordinates.
This setting brings two benefits.
On the one hand, it eliminates the variance of body sizes of subjects in datasets.
On the other hand, it ensures that the lengths of human limbs in each generated sequence are consistent.

Action instances (whether at input or output) in our method are $T \times C$ skeleton action sequences.
$T$ indicates the temporal length (unified to 32 frames long on both two datasets) and $C$ is the dimension of a 3D human pose in one frame (normally $C = number \ of \ limbs \times 3$).
1D convolutions are performed in our various networks.
All GAN-based models in the following experiments are trained under WGAN-GP~\cite{wgangp}.

\subsection{Comparison with GAN-based Data Augmentation}

\setlength{\tabcolsep}{4pt}
\begin{table}[b]
\begin{center}
\caption{
Quantitative comparison of data augmentation effects between CSGN and PE.
}
\label{table:comparison}
\begin{tabular}{c c c c c}
    \hline
    \noalign{\smallskip}
     & \multicolumn{2}{c}{UTD-MHAD} & \multicolumn{2}{c}{AID} \\
    \noalign{\smallskip}
    \makebox[30mm]{Data Augmentation} &
    \makebox[15mm]{IAT-test} &
    \makebox[15mm]{IAT-train} &
    \makebox[15mm]{IAT-test} &
    \makebox[15mm]{IAT-train} \\
    \noalign{\smallskip}
    \hline
    \noalign{\smallskip}
    -- & 85.32 & 53.92 & 87.29 & 51.17 \\
    \noalign{\smallskip}
    CSGN~\cite{CSGN} & 87.86 & 58.97 & 89.96 & 68.82 \\
    \noalign{\smallskip}
    \textbf{PE (Ours)} & \textbf{91.03} & \textbf{64.94} & \textbf{90.69} & \textbf{75.65} \\
    \noalign{\smallskip}
    \hline
\end{tabular}
\end{center}
\end{table}
\setlength{\tabcolsep}{1.4pt}

 As discussed in Sec. 1 and 2.1, GAN-based augmentation methods for classification and other specified tasks can not be applied to our task.
Therefore, training an unconditional GAN for directly generating action pairs is left as the only choice for GAN-based data augmentation.
We select CSGN~\cite{CSGN}, which is promising to generate high-quality human action sequences unconditionally.
A comparison of data augmentation effects between our PE method and this method is shown in Table.~\ref{table:comparison}.

Learning without augmentation gives generation results that are acceptable from reality and precision (a 85.32/87.29 IAT-test score), but extremely disappointing in diversity (a 53.92/51.17 IAT-train score).
For augmentation, a CSGN network is first trained to model the distribution of paired action data.
Then we mix generated action pairs with existing data to train our Act2Act network.
This method benefits the learning of the task followed, especially visible from a significant increase in IAT-train score.
However, it still lags behind our method 3.17/0.73 and 5.97/6.83 respectively in two metrics.
We examine generated actions from CSGN and find them to be realistic but not diverse enough, thus provide limited modes for augmentation.
Such results keep in line with the fact that GAN-based methods need large-scale training data to ensure multi-modal generation quality.
As a comparison, our PE method is more friendly to this small-scale data.
Considerable improvements in diversity of generated action responses reflect similar improvements brought by PE in diversity of paired training data.

\subsection{Ablation Study}

\subsubsection{Embedding Space.}

\begin{figure}[t]
\centering
\includegraphics[width=120mm]{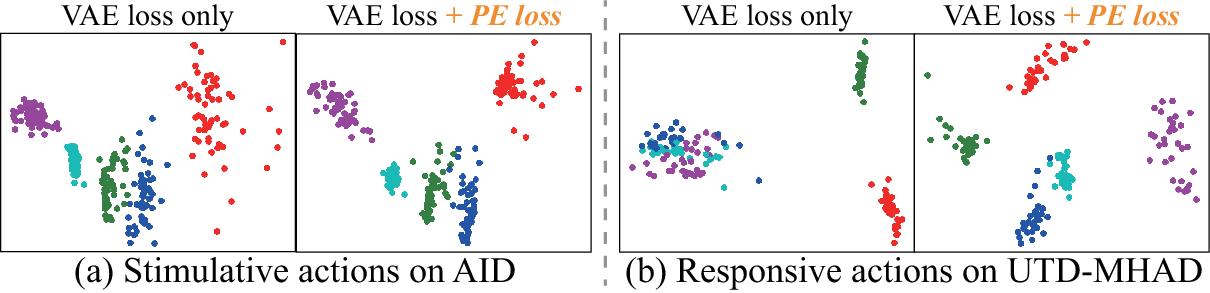}
\caption{
Action embeddings projected by PVAEs trained with VAE loss only and with PE loss also.
}
\label{fig:cluster}
\end{figure}

Fig. 5 visualizes the distribution of actions in the embedding space, projected by PVAEs trained with/without PE loss.
Groundtruth category labels are utilized to color data points for comparison.
As can be seen, additional PE loss brings much better clustering effects in both gatherings within categories (especially in Fig.~\ref{fig:cluster}(a)) and distances between categories (in Fig.~\ref{fig:cluster}(b)).

We analyze two critical hyper-parameters affecting PE data augmentation: the scale factor $s$ and the dimension of embedding space $\mathbb{L}$.
Augmentation effects reflected in $NC$ matrices are evaluated from effectiveness $F$ and reliability $R$.
Specifically, $F$ is represented as the probability that each sample is replaced by others to form new pairs, i.e.,
\begin{equation}
    F = \frac{1}{N} \sum_{i=1}^N \sum_{j=1}^N 1(i \neq j) \cdot NC(i, j) .
\end{equation}
Meanwhile, we import groundtruth category labels to calculate the probability of category unchanged after replacement as $R$, i.e.,
\begin{equation}
    R = \frac{1}{N} \sum_{i=1}^N \sum_{j=1}^N 1(cat^{(i)} = cat^{(j)}) \cdot NC(i, j) ,
\end{equation}
where $cat$ is the category of action.

\begin{figure}[t]
\centering
\includegraphics[width=120mm]{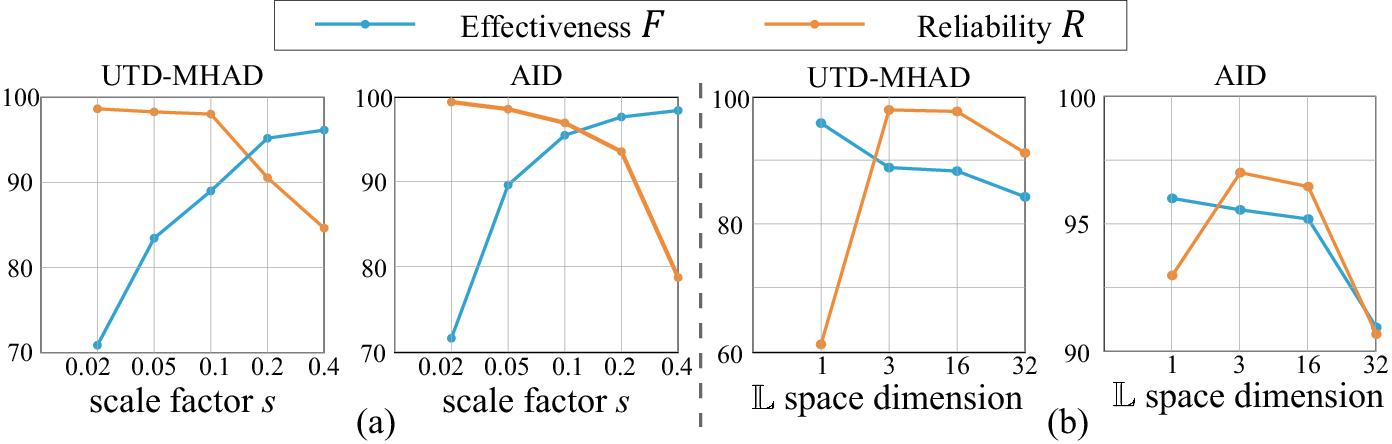}
\caption{
Data augmentation effects with different (a) scale factors and (b) dimensions of $\mathbb{L}$ space.
}
\label{fig:hyperparam}
\end{figure}

As the neighborhood range controlled by $s$ expands, the effectiveness of PE data augmentation increases while the reliability decreases.
Fig.~\ref{fig:hyperparam}(a) suggests $s = 0.1$ to be the equilibrium point of $F$ and $R$ on both two datasets.
Changes brought by different dimensions of $\mathbb{L}$ space are more complicated.
As Fig.~\ref{fig:hyperparam}(b) shows, when $\mathbb{L}$ is a 1-d space, learning PVAEs to cluster actions in it can be difficult.
The low reliability reflects relatively weak clustering effects at this time.
Then the subtle difference between 3-d and 16-d suggests a very flexible selection range for a reasonable embedding space dimension.
When the dimension further increases, augmentation effects start to corrupt, mostly due to the imbalance between PE loss and VAE loss during training PVAEs.

\subsubsection{Comparison with Label-Given Methods.}

Here experiments are conducted in label-given situations to give an upper bound of performance of our method:

\setlength{\tabcolsep}{4pt}
\begin{table}[t]
\begin{center}
\caption{
Quantitative comparison of generation effects between our proposed method and methods in label-given situations.
}
\label{table:ablation}
\begin{tabular}{c c c c c c}
    \hline
    \noalign{\smallskip}
     & & \multicolumn{2}{c}{UTD-MHAD} & \multicolumn{2}{c}{AID} \\
    \noalign{\smallskip}
    \makebox[15mm]{Data} &
    \makebox[15mm]{Label-given} &
    \makebox[15mm]{IAT-test} &
    \makebox[15mm]{IAT-train} &
    \makebox[15mm]{IAT-test} &
    \makebox[15mm]{IAT-train} \\
    \noalign{\smallskip}
    \hline
    \noalign{\smallskip}
    Original & $\times$ & 85.32 & 53.92 & 87.29 & 51.17 \\
    \noalign{\smallskip}
    \textbf{PE aug.} & $\times$ & 91.03 & 64.94 & 90.69 & 75.65 \\
    \noalign{\smallskip}
    \hline
    \noalign{\smallskip}
    Re-assign & \checkmark & 90.97 & 68.93 & 93.05 & 82.15 \\
    \noalign{\smallskip}
    Split & \checkmark & 91.35 & 71.64 & 95.04 & 85.89 \\
    \noalign{\smallskip}
    \hline
\end{tabular}
\end{center}
\end{table}
\setlength{\tabcolsep}{1.4pt}

\textbf{1) Re-assign:} Actions are re-assigned into new pairs according to groundtruth labels.
All paired relationships conforming to defined interaction rules are exhausted for the training of Act2Act.

\textbf{2) Split:} The network is explicitly split into two parts: a classification part for stimulative actions and a category-conditional generation part for responsive actions.
The two parts are independently trained with category labels given and connected in series during inference.

As Table.~\ref{table:ablation} shows, methods augmented by PE is very close to label-given methods in performance, compared to the original baseline.
With category labels given, we can attain more satisfactory generation results.


\subsection{Qualitative Evaluation}

\begin{figure}[h!]
\centering
\includegraphics[width=120mm]{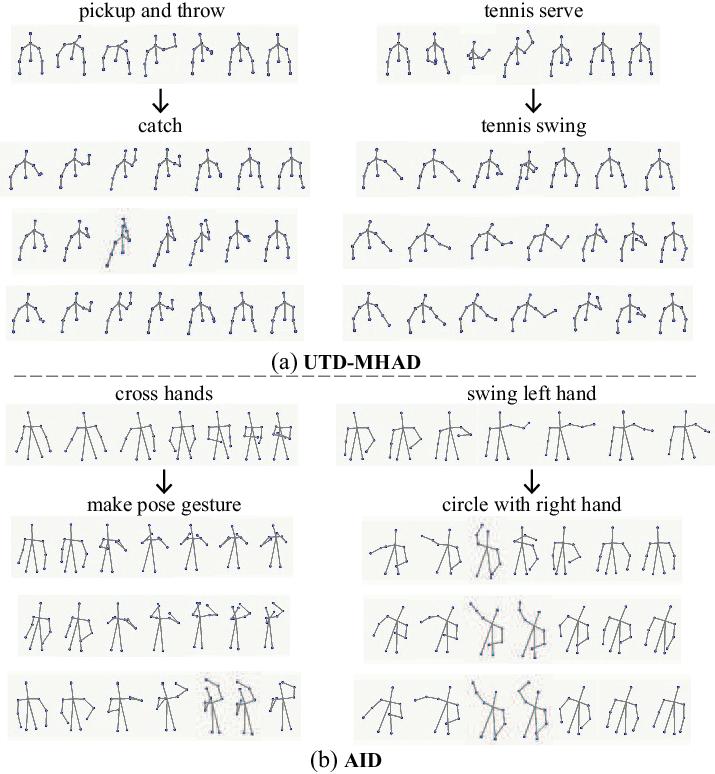}
\caption{
Examples of generation on two datasets.
For each example, the given stimulative action and generated responses corresponding to three random noise vectors are shown.
Visualized actions are meanly sampled from 32-frame sequences.
}
\label{fig:gen}
\end{figure}

Generated responses conditioned on some stimulative actions are shown in Fig.~\ref{fig:gen}.
Three fixed random noise vectors $z_1$, $z_2$ and $z_3$ are involved in each generation.
We first examine how responsive actions are generated with a fixed stimulation and different random noise vectors.
It is surprising to note that given the same stimulative action, generated responses are various due to randomness from $z$.
Such variety of actions manifests in several aspects like pose, movement speed and range.

Secondly, we examine how responsive actions are generated with a fixed random noise vector and different stimulations.
As can be seen, all generated responses belong to respective categories expected by interaction rules.
This indicates that within our method, latent code $c$ in Act2Act precisely controls semantic translation.
In addition, human-like vividity shown in these generated actions is impressive.
Overall, qualitative evaluation further verifies the effectiveness of our method in meeting all three generation goals.

\section{Conclusion and Future Work}

In this paper, we specify a novel task to learn end-to-end action interaction and propose a PE data augmentation method to enable learning with small-scale unlabeled data.
Another Act2Act network learns from augmented data.
Two new metrics are also specially designed to evaluate methods on our task from generation goals of reality, precision and diversity.
Our PE method manages to augment paired action data significantly and reliably.
Experimental results show its superiority to baseline and other GAN-based augmentation methods, approximating the performance of label-given methods.
Given impressively high-quality action responses generated, our work shows broad application prospects in action interaction.
We also hope our PE method to enlighten other unsupervised learning tasks with weak information like paired relationships in our task.

In the future, we plan to advance research in two directions.
On the one hand, we aim to transfer our method to other output forms, like low-level control parameters of a robot platform.
Thus generated responses can be directly applied in robot control.
On the other hand, we expect to learn from unsegmented long interaction sequences instead of segmented clips to further simplify the data collection of our task.



\bibliographystyle{splncs}
\bibliography{my_paper}

\end{document}